%% file: paper.tex
\documentclass[conference]{IEEEtran}
\IEEEoverridecommandlockouts

\makeatletter
\def\ps@headings{%
\def\@oddhead{\mbox{}\scriptsize\rightmark \hfil \thepage}%
\def\@evenhead{\scriptsize\thepage \hfil \leftmark\mbox{}}%
\def\@oddfoot{}%
\def\@evenfoot{}}
\makeatother
\usepackage{algorithm}
\usepackage{algorithmicx}
\usepackage{algpseudocode}
\usepackage{amsmath}
\usepackage{graphicx}
\usepackage{subfigure}
\usepackage{times}
\usepackage[hyphens]{url}
\usepackage{color}
\usepackage{multirow}
\usepackage{textcomp}
\usepackage{threeparttable}
\usepackage{supertabular}
\usepackage{amssymb}
\usepackage{amsfonts}
\usepackage{booktabs}
\usepackage{epstopdf}
\usepackage{xspace}
\usepackage{enumitem}
\usepackage{cite}
\usepackage{array}
\usepackage{graphicx}

\usepackage{fancyhdr}

\newcommand{\MIMIC}{{\sc MIMIC III}\xspace}
\newcommand{\FineEHR}{{\sc FineEHR}\xspace}

\renewcommand{\paragraph}[1]{\smallskip\noindent {\bf #1}}

\begin{document}

\title{FineEHR: Refine Clinical Note Representations to Improve Mortality Prediction}

\author{\IEEEauthorblockN{1\textsuperscript{st} Jun Wu$^*$}
\IEEEauthorblockA{
\textit{Georgia Institute of Technology}\\
Atlanta, United States \\
jwu772@gatech.edu}
\and
\IEEEauthorblockN{2\textsuperscript{nd} Xuesong Ye}
\IEEEauthorblockA{
\textit{Trine University}\\
Phoenix, United States \\
xye221@my.trine.edu}
\and
\IEEEauthorblockN{3\textsuperscript{rd} Chengjie Mou}
\IEEEauthorblockA{
\textit{Trine University}\\
Phoenix, United States \\
cmou22@my.trine.edu}
\and
\IEEEauthorblockN{4\textsuperscript{th} Weinan Dai}
\IEEEauthorblockA{
\textit{Trine University}\\
Phoenix, United States \\
wdai22@my.trine.edu}
}

\maketitle

\thispagestyle{fancy}
\fancyhead{}
\lhead{}
\lfoot{979-8-3503-3698-6/23/\$31.00 ©2023 IEEE \hfill}
\cfoot{}
\rfoot{}

\input{abstract}
\input{intro}

\input{dataset}
\input{design}

\input{evaluation}
\input{conclusions}

\bibliographystyle{IEEEtran}
\bibliography{paper}

\end{document}

%% file: abstract.tex
\begin{abstract}

Monitoring the health status of patients in the Intensive Care Unit (ICU) is a critical aspect of providing superior care and treatment. The availability of large-scale electronic health records (EHR) provides machine learning models with an abundance of clinical text and vital sign data, enabling them to make highly accurate predictions. Despite the emergence of advanced Natural Language Processing (NLP) algorithms for clinical note analysis, the complex textual structure and noise present in raw clinical data have posed significant challenges. Coarse embedding approaches without domain-specific refinement have limited the accuracy of these algorithms. To address this issue, we propose \FineEHR, a system that utilizes two representation learning techniques, namely metric learning and fine-tuning, to refine clinical note embeddings, while leveraging the intrinsic correlations among different health statuses and note categories. We evaluate the performance of \FineEHR using two metrics, namely Area Under the Curve (AUC) and AUC-PR, on a real-world MIMIC III dataset. Our experimental results demonstrate that both refinement approaches improve prediction accuracy, and their combination yields the best results. Moreover, our proposed method outperforms prior works, with an AUC improvement of over 10\%, achieving an average AUC of 96.04\% and an average AUC-PR of 96.48\% across various classifiers.

\end{abstract}


%% file: intro.tex
\section{Introduction}
\label{sec:intro}

\subsection{Background and Motivation}
Mortality prediction, which aims to forecast the probability of patient death, is crucial for timely and improved risk assessment in clinical services. Intensive care units (ICUs) produce and record massive structured or unstructured clinical data, increasing the potential for machine learning models to predict and analyze automatically instead of relying on specialists. Many researchers used advanced natural language processing algorithms to learn document embeddings from textual data, then make prediction models for downstream tasks. Clinical notes tend to be fragmented and have different writing styles due to individual differences between recorders, which are very noisy for machine learning models and limit the performance of prediction tasks overall. Our work investigates representation learning techniques to overcome this challenge. 

\subsection{Mortality prediction via Electronic Health Records}
Previous works in mortality prediction can be categorized into three approaches: the \emph{time series-based approach}, the \emph{texts-based approach}, and the \emph{hybrid approach}.

\label{sec:related}
\subsubsection{Time series-based}
\label{subsec:mortality}
These works use clinical time series data to analyze and predict the health status of patients, including vital signs (e.g., heart rate, blood pressure) and laboratory test results (e.g., electrolytes, blood glucose). Harutyunyan \emph{et al.} \cite{harutyunyan2019multitask} proposed four prediction tasks on the \MIMIC dataset, including \emph{In-hospital Mortality}, \emph{Decompensation}, \emph{Length of Stay}, and \emph{Phenotyping}, and also explored the effects of multitask learning using clinical time series data on all tasks. Facing the challenge of missing data, Che \emph{et al.} \cite{che2018recurrent} used an RNN-based approach to interpolate missing data in multivariate clinical time series data, significantly improving prediction accuracy. Narayan Shukla \emph{et al.} \cite{shukla2021multi} proposed a multi-attention based mechanism to solve the problem of irregularly sampled clinical time series data, improving prediction accuracy on tasks.

\subsubsection{Texts-based} Clinical notes are textual data recorded by healthcare providers. This approach models the semantic information of clinical notes to reflect the health status of patients and then make essential predictions like mortality, readmission, and stays-days. Clinical note categories include nursing notes, discharge summaries, and physician notes, which are recorded by healthcare professionals during patient hospital stays. Boag \emph{et al.} \cite{boag2018s} proposed extracting critical information and values from clinical notes to predict mortality. Liu \emph{et al.} \cite{liu2019knowledge} introduced a knowledge-aware deep dual network to utilize external medical knowledge to guide mortality prediction. Huang \emph{et al.} \cite{huang2019clinicalbert} designed ClinicalBERT, a domain-specific adaptation of the pre-trained BERT model, optimizing semantic embeddings of clinical notes to improve the prediction accuracy of hospital readmissions. Jin \emph{et al.} \cite{jin2018improving} using named entities recognition to discover important words in clinical notes to enhance hospital mortality prediction.

\subsubsection{Hybrid} These researches combine multiple data types like clinical textual, time series, image to profile a patient's health status, enabling more accurate predictions of clinical tasks. Ghorbani \emph{et al.} \cite{ghorbani2020new} and Khadanga \emph{et al.} \cite{khadanga2019using} designed a hybrid predictive model combining patient health signs in clinical time series and notes and performed well in mortality prediction on imbalanced datasets. Deznabi \emph{et al.} \cite{deznabi2021predicting} combined clinical notes and time series data to create a more comprehensive representation of patients' health status for accurate mortality prediction. Zheng \emph{et al.} \cite{liu2020heterogeneous} proposed a novel heterogeneous graph approach to model the correlation between patients and diseases to make better prediction performance. Fan \emph{et al.} \cite{wang2021topotxr} and \cite{wang2020topogan} designed attention neural network to extract tissue topological structures from magnetic resonance imaging (MRI) data for predicting breast cancer.

\subsection{Representation Generation and Refining}
Embedding unstructured data across various modals typically involves two key steps: \emph{generating embedding} and \emph{refining embedding}. We first introduce some classic algorithms and advanced application works of embedding generation across various data sources, including \emph{text}, \emph{image}, and \emph{multisource data}. Then we discuss embedding refining approaches for improving downstream tasks' performance, including \emph{fine-tuning}, \emph{metric method}, \emph{reweighing and sampling}.

\subsubsection{Embedding Generation} This field targets mapping high-dimensional data into a numerical vector space. In the field of \textbf{natural language processing (NLP)}, the generation operation represents words, phrases, and sentences into numerical vectors. Popular methods include Word2Vec \cite{mikolov2013distributed}, GloVe \cite{pennington2014glove}, and Transformer \cite{vaswani2017attention}, which leverage various techniques such as neural networks, co-occurrence statistics, and contextual language modeling to capture the semantic and syntactic properties of textual data.

\textbf{Image and video data} are widely collected in hospitals. Image data are represented to high-dimensional space before make prediction. The most famous work is AlexNet \cite{krizhevsky2017imagenet}, which learns hierarchical feature representations from raw image data. Currently, DeepMTL Pro \cite{zhan2022deepmtl} proposed a novel image-to-image translation framework via image representing and understanding. Many deep learning and machine learning models represent image data to diagnose diseases. Syed Muhammad \emph{et al.} \cite{anwar2018medical} is a survey to introduce how convolutional neural network (CNN) works for medical image analysis. Yuli \emph{et al.} \cite{wang2023investigation} utilized probability maps to represent magnetic resonance images (MRIs) for brain disorder analysis. Liangliang \emph{et al.} \cite{ren2018collaborative} proposed a deep reinforcement learning model to detect and track multiple objections in video frames. Xueshen \emph{et al.} \cite{li2023detectingmeasure} and \cite{li2023detecting} designed spatial-temporal deep learning algorithms to detect and measure gastric diseases. 

Jointly learning from textual, image and time series data (\textbf{multi-source data}) and make representation is challenge and essential for tasks in the real-world environment. Current researches made some breakthroughs. Yuxin \emph{et al.} \cite{tian2023fashion} designed a multi-modal transformer-based architecture and applied it in e-commerce applications. ALBEF \cite{li2021align} aligned text and image embeddings to do text-image matching. GL-RG \cite{yan2022gl} can generate text with similar semantic information from visual signals. CellPAD \cite{wu2018cellpad} combined temporal signals and local patterns in time series to predict network anomalies. Xiaoling \emph{et al.} \cite{luo2022multisource} combined time-series signs and video to estimate vehicle traffic. BotShape \cite{wu2023botshape} represented social account behavioral logs into sequence embeddings for bot detection. Shujie \emph{et al.} \cite{han2019robust} combined textual system logs and hard disk time series to predict disk failure in data center.

\subsubsection{Embedding Refining} This technique improves the quality of initially generated embeddings and represents them as new embeddings. \emph{Metric learning} and \emph{model fine-tuning} are two main refining technqiues.

\textbf{Metric learning} was first applied in computer vision by adjusting the distance between positive and negative samples for better prediction task performance. Bertinetto \emph{et al.} \cite{bertinetto2016fully} proposed a fully-convolutional siamese network architecture for object tracking and presented accurate performance across various scenarios. Zhuoyi \emph{et al.} \cite{wang2020adaptive} and \cite{gao2019sim} target to build a robust and general metric-learning solution to deal with open-world problems. Our previous work BotTriNet \cite{wu2023bottrinet} applied the triplet network to textual embedding representation, vastly improving social bots detection accuracy.

The most efficient application of \textbf{fine-tuning} technique is BERT \cite{devlin2018bert}, fine-tuning parameters of the pre-trained model, showing outperformance on various NLP tasks. ULMFiT \cite{howard2018universal} applied transfer learning method for fine-turning a language model. 
Statistical re-weighting is a common strategy to solve medical data imbalance problem. However, PriMeD \cite{liu2022mitigating} abandoned the solution proposed an intelligent deep learning method via auto-encoder.

%% file: dataset.tex
\section{Dataset}
\label{sec:dataset}
\begin{table}[h]
\centering
\caption{Top-frequent Clinical Note Categories in \MIMIC}
\label{tab:notecate}
\vspace{-6pt}
\begin{tabular}{|p{0.45in}|l|p{2.2in}|}
\hline
{\bf Category} & {\bf Ratio} & {\bf Descriptions}\\
\hline
Discharge summary & 97.57\% & A summary of a patient diagnosis, treatment during discharge (has no mortality status). \\
\hline
ECG & 87.58\% & A text that records the electrical activity of the heart to help diagnose heart conditions. \\
\hline
Radiology & 82.95\% & A medical imaging record such as X-rays, CT scans, and MRI. \\
\hline
Nursing / other & 51.39\% & Notes of general nursing care or other issues. \\
\hline
Echo & 45.82\% & A textual record of imaging test using sound waves to create pictures of the heart. \\
\hline
Nursing & 19.26\% & A note related specific nursing care. \\
\hline
Physician & 19.16\% & A note written by a physician or health carers. \\
\hline
\end{tabular}
\end{table}

\MIMIC is a real-world clinical and health-related dataset, which includes thousands of ICU patient admission records in a medical center between 2001 and 2012. The dataset includes extensive electronic health records (EHR), including medications, laboratory results, vital signs, and diagnosis codes. We focus on the mortality prediction task. The mortality label is positive if a patient dies during their hospital stay. We use clinical notes to extract semantic embeddings as \textbf{features}. Clinical notes refer to text information recorded by healthcare professionals, including different types such as nursing notes, operative notes, progress notes, and others, each containing different kinds of information. Table~\ref{tab:notecate} provides descriptions of clinical note categories whose proportion is over 15\%. The ratio refers to the percentage of the corresponding note category that appears as the total number of hospital admissions.

%% file: design.tex
\section{Design}
\begin{figure}[htb]
\centering
\includegraphics[width=\linewidth]{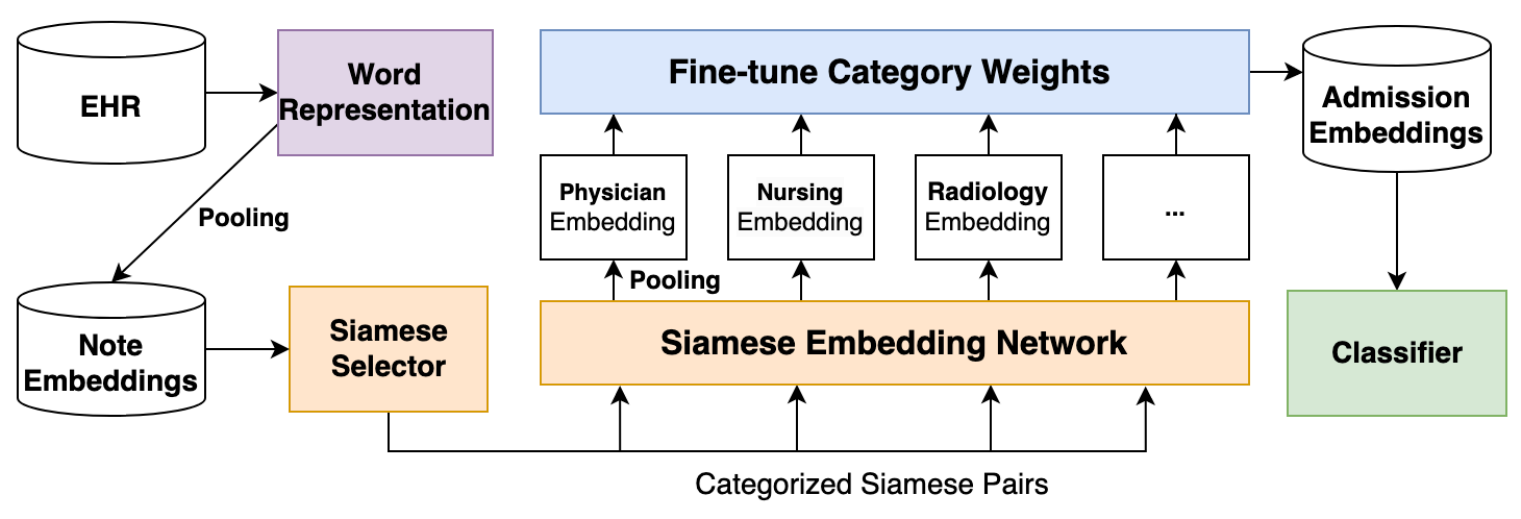}
\vspace{-6pt}
\caption{\FineEHR Architecture.}
\label{fig:arch}
\end{figure}
\label{fig:design}

We present \FineEHR, a system for refining clinical note representations for accurate mortality prediction. It takes clinical notes from ICU patients' electronic health records (EHR) as input and predicts whether a patient will die in the future. Figure~\ref{fig:arch} shows the architecture of our system.

\FineEHR first represents the original textual notes in semantic spaces via the word embedding technique and then generates note-level, category-level, and admission-level embeddings through the pooling method. To refine the above embeddings for better mortality prediction, considering the textual structure differences among various note categories, \FineEHR utilizes two representation learning techniques, including \textbf{metric learning} and \textbf{fine-tuning} at the middle stages. Specifically, it first uses a Siamese Network \cite{bromley1993signature} to adjust the distances between positive and negative note embeddings for each category. After producing category-level embeddings through average pooling of adjusted note embeddings, \FineEHR learns each category's weights (importance) by fine-tuning on training instances and then aggregating all category-level embeddings into the unique admission-level embedding for each patient via pooling according to corresponding weights. Finally, the tuned admission-level embeddings support various classifiers for mortality prediction.

\subsection{Word Representation}
Word embedding is a fundamental technique in the field of NLP. It transforms words into lower dimensions with a vector representation as the base material for downstream NLP modeling. The widely applied word representation algorithms include Word2Vec \cite{mikolov2013efficient}, GloVe \cite{pennington2014glove}, and BERT \cite{vaswani2017attention}. \FineEHR utilizes Word2Vec to generate clinical word embeddings based on the textual \MIMIC notes. To avoid transforming label information to test samples, we restrictively used only texts in train samples to fit and generate word embeddings. \FineEHR selected Word2Vec as the word representation approach because it outperformed GloVe in the performance of our task, and its embedding generation speed was faster than the pre-trained model called BioBERT \cite{lee2020biobert}.

The detailed implementation first split texts into sentences, then split each sentence into continuous word sequences (called \textbf{tokenization}). Then, it inputs word sequences into the Word2Vec model, which utilizes the surrounding words to make word presentations. The structure of EHR notes is irregular, and we did extra \textbf{text preprocessing} for original texts before tokenization. In detail, we use a newline as the splitting point, even if it looks incomplete, except for standard ending punctuation between sentences. We also convert all capital letters to lowercase before tokenization to handle multiple versions of the same medical vocabulary.

\subsection{Refine Note Embedding via Siamese Network}
\label{subsec:pool}

\FineEHR applies average pooling to generate note-level embeddings for each clinical note. Average pooling is a simple but effective approach in sentence embedding generation and downstream classification tasks \cite{kalchbrenner2014convolutional}. The dimension size of the note-level embedding and word embedding is the same after directly averaging values from all words in each dimension.

\FineEHR chose Siamese Network to refine the original note embedding, inspired by Sentence BERT \cite{reimers2019sentence}, which used a metric approach to represent sentence embedding to enhance the performance of downstream text tasks and outperformed the original sentence embeddings of BERT model. \FineEHR did a domain-specific adaption of the Siamese Network after observing the inner similarity between same-category notes and the extensive content diversity of notes among different categories. We respectively introduce three core modules, including \emph{Siamese Selector}, \emph{Siamese Embedding Network}, and \emph{Siamese Loss}.

\subsubsection{Siamese Selector}

Siamese Selector refers to the algorithm for selecting Siamese Pairs. The overall training objective of the Siamese Network is to bring samples closer if they have the same label and move samples far from each other if their labels are different. In Siamese Networks, the format of an input instance is a pair during iterations. In detail, it randomly selects two samples of note embeddings at each time (called the \emph{Anchor} and \emph{Contrast}, collectively called a \textbf{Siamese Pair}). Specifically, if the contrast sample shares the same label as the anchor sample, the Pair is considered a \textbf{positive instance}; otherwise, it is a \textbf{negative instance}. The positive or negative correlation is passed to the Siamese Network to calculate the loss function.

To refine clinical representations better, \FineEHR made a domain-specific adjustment to the Siamese Selector setting. Specifically, \FineEHR divided samples of note embeddings into various groups based on their category. Pairs are randomly selected only within each group (category), with the correlation of labels determining whether they are positive or negative. 

As illustrated in Table~\ref{tab:notecate}, the content of different clinical note categories varies significantly, resulting in natural differences among note embeddings in the semantic space. Consequently, applying metric learning across different types of notes is deemed meaningless. We discuss the consequences of the two methods of not using a categorized manner. If two notes with the same mortality label but different categories are forced to be closer in the distance, their natural distributions (embeddings) may interfere with each other; if their original mortality statuses differ, pushing them away from each other is superfluous because they already distributed separately in semantic space. In conclusion, the categorized Siamese Selector enhances the precision of the embeddings and decreases the number of iterating epochs required.

\subsubsection{Siamese Embedding Network}

Siamese Embedding Network is a representation learning network that can refine embedding by transforming raw embeddings into tuned embeddings. The network structure can be simple or complex. \FineEHR chooses a symmetrical multilayer perceptron structure (MLP) as the representation network, where the input and output layer dimensions are equal, and the dimensions of each layer increase first and decrease then. We did not spend more time designing complex network structures because we found that the effect of MLP is remarkable enough.

\subsubsection{Siamese Loss}

The loss function in a Siamese Network optimizes the embedding space by minimizing the distance between similar samples (positive pairs) and between dissimilar samples (negative pairs). The goal is to improve the model's ability to differentiate between distinct classes or categories in the latent space.

A standard implementation of the Siamese Network loss function is the contrastive loss. It operates by calculating the Euclidean distance between the embeddings of the anchor and contrast samples. In detail, the contrastive loss function has two parts: one targeting positive and the other optimizing negative Siamese pairs.

Siamese Network adjusts the parameters of the Embedding Network via the gradient descent method to minimize the following contrastive loss function: \\

$Loss = Y * D^2 + (1-Y) * max(margin - D, 0)^2 $
\\
\\
With $Y$ the label of Siamese Pair (positive(Y=1) or negative(Y=0)), $D$ the Euclidean distance between the two raw embeddings of anchor and contrast samples in the pair. $margin$ is a hyper-parameter to ensure that anchor and contrast samples with different labels are separated by a minimum distance in the embedding space.

\subsection{Refine Admission Embedding via Fine-tuning}

\begin{figure}[htb]
\centering
\includegraphics[scale=0.5]{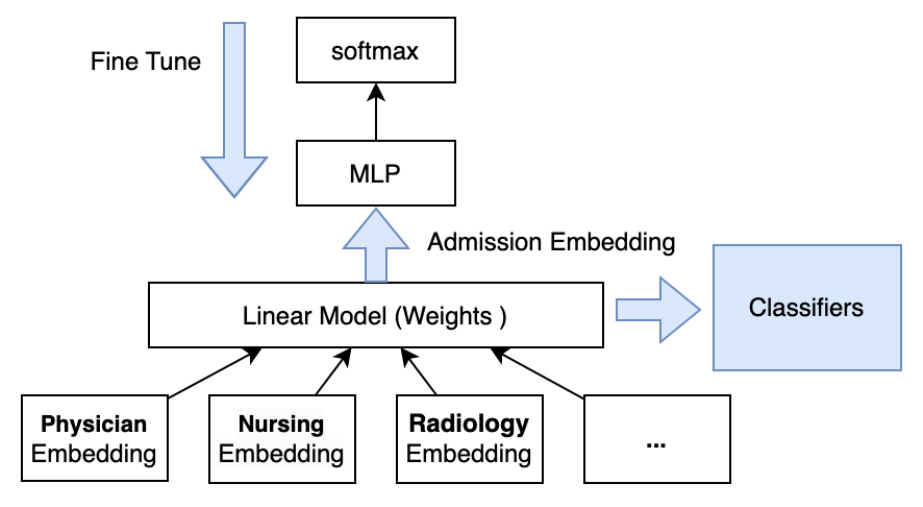}
\vspace{-6pt}
\caption{Category Weights Fine-tuning.}
\label{fig:rewieght}
\end{figure}

As we mentioned in Section~\ref{sec:dataset}, the definitions and ratios of note categories are different. We naturally consider the contributions of different clinical note categories to be various. For example, the Physician category contains more professional descriptions of diseases and symptoms, while the General category only records unimportant information like meals and schedules. Based on the observation and understanding of clinical notes,  \FineEHR defined fine-tuning approach based on categories to allocate different embeddings with different importance. Figure~\ref{fig:rewieght} shows the model design.

First, our system calculates the \textbf{category-level embeddings} for each note category by average pooling all notes with the corresponding category in one patient's admission. Then it sets a linear representation model to learn weights for each category and represent multiple category-level embeddings into only one embedding (called refined admission-level embedding). The system optimizes the linear model weights based on the classification accuracy of the mortality prediction task. The system uses another linear MLP as the classifier to fit the correlation between the admission-level embedding and the mortality label. After iterations, the model outputs the parameters in the first linear model as the optimized weights and output layer embedding as the \textbf{admission-level} embedding for various classifiers to make predictions.

\FineEHR outputs refined admission-layer embedding for downstream classifiers to fit training data. In other words, its main target is to learn a better representation of clinical notes, but it leaves the flexibility for users to choose different classification algorithms (e.g., Random Forest, Logistic Regression, or Deep Network). The sub section~\ref{subsubsec:clf} gives the performance across different classifier based on the refined admission-level embeddings.

%% file: evaluation.tex
\section{Evaluation}
\label{sec:evaluation}

\subsection{Ground Truth and Performance Metrics}

\begin{table*}[htbp]
  \centering
  \caption{AUC and AUC-PR scores during different settings in \FineEHR}
  \vspace{-6pt}
    \begin{tabular}{|c|c|c|c|c|c|c|c|c|}
    \hline
          & \multicolumn{2}{c|}{\textbf{RF}} & \multicolumn{2}{c|}{\textbf{LR}} & \multicolumn{2}{c|}{\textbf{MLP}} & \multicolumn{2}{c|}{\textbf{GBDT}} \\
    \hline
    \multicolumn{1}{|l|}{\textbf{Approach}} & \textbf{AUC} & \textbf{AUC-PR} & \textbf{AUC} & \textbf{AUC-PR} & \textbf{AUC} & \textbf{AUC-PR} & \textbf{AUC} & \textbf{AUC-PR} \\
    \hline
    Baseline & 86.73\% & 88.72\% & 89.52\% & 90.75\% & 86.41\% & 87.35\% & 87.30\% & 88.80\% \\
    \hline
    Only Metric & 94.10\% & 95.01\% & 92.69\% & 93.84\% & 91.05\% & 93.15\% & 93.53\% & 93.59\% \\
    \hline
    Only Weight & 95.09\% & 95.36\% & \multicolumn{1}{c}{94.35\%} & 95.02\% & 94.41\% & 93.90\% & 96.02\% & 96.50\% \\
    \hline
    FineEHR & \textbf{97.18\%} & \textbf{97.77\%} & \textbf{95.18\%} & \textbf{96.08\%} & \textbf{94.91\%} & \textbf{95.35\%} & \textbf{96.88\%} & \textbf{96.72\%} \\
    \hline
    \end{tabular}%
  \label{tab:acc}%
\end{table*}%

\textbf{Ground-truth} is a helpful tool for evaluating the prediction performance with labeled instances. We conducted a ground-truth evaluation through the \MIMIC dataset to evaluate \FineEHR. In the data, the identification of hospital admission is \emph{HADM\_ID}. The id correlates with an attribute \emph{HOSPITAL\_EXPIRE\_FLAG} (called flag), recording the label of mortality and all related clinical notes during this admission. The flag equals one stand for the mortality status is actual, and we set the hospital admission as a positive instance. Otherwise, the admission is a negative instance.

The raw dataset is imbalanced. We equaled the numbers of positive and negative instances for a better performance evaluation via random downsampling on the negative samples. All the negative instances and sampled positive instances were randomly split into two groups called \textbf{training set} and \textbf{testing set} (testing occupying 20 percent).

\textbf{Performance metrics} are essential for comparing different models or approaches. We used \emph{AUC} and \emph{AUC-PR} as evaluation metrics. \textbf{AUC} is the area under the Receiver Operating Characteristic (ROC) curve, which plots the True Positive Rate (TPR) against the False Positive Rate (FPR). A higher AUC value indicates that the model performs better in discriminating between the positive and negative classes across various threshold settings. \textbf{AUC-PR} is the area under the precision-recall curve. The higher it is, the more mortality instances would be accurately predicted, and the fewer false alarms would appear.

\subsection{Accuracy Improvements via Refining Approaches}
\label{subsubsec:clf}
To present the effectiveness of two refining approaches, including Siamese Network and Category Weights Fine-tuning, we conducted baseline experiments using raw embeddings without any refining process for better comparison.

\textbf{Baseline Approach:} In subsection~\ref{subsec:pool}, \FineEHR obtains raw note-level embeddings via average pooling. The baseline approach then directly computes the admission-level embedding through an average pooling of all related note-level embeddings.

\textbf{Only Metric:} We train a Siamese Network to refine the raw note-level embeddings separately in each category (using only instances in the training set). After refinement, \FineEHR directly computes the admission-level embedding by averaging all the refined note-level embeddings.

\textbf{Only Weight:} Without Siamese Network refinement, \FineEHR uses raw note-level embeddings to fine-tune the category weights in a linear model, then pools them into an admission-level embedding through learned weights for different categories.

Four settings (including the complete version of \FineEHR) generate different admission-level embeddings. We use four well-known classifiers to make mortality predictions and evaluate the predicted labels using the testing set. RF refers to Random Forest. LR refers to Logistic Regression. MLP refers to Multi-layer Perceptron. GBDT refers to Gradient-Boosted Decision Trees. For a fair comparison, all classifiers are under default parameter settings without parameter optimization for the four different settings.

Table~\ref{tab:acc} shows that both the Metric and Weight approaches significantly improve the prediction accuracy. Moreover, when combining the two refinement approaches, \FineEHR achieves the highest performance. These results demonstrate that the two refining approaches are very effective, independently or combined.

Table~\ref{tab:acc} shows the AUC and AUC-PR under different settings in \FineEHR. It shows that both the Metric and Weight approaches significantly improve the prediction accuracy. Moreover, when combining the two refinement approaches, \FineEHR achieves the highest performance. These results demonstrate that the two approaches are effective, whether independently or combined.

\subsection{Compared With Previous Works}
\begin{table}[htbp]
  \centering
  \caption{Compare AUC with previous works}
  \vspace{-6pt}
    \begin{tabular}{|p{1.2in}|>{\centering\arraybackslash}p{0.8in}|>{\centering\arraybackslash}p{0.6in}|}
    \hline
    \textbf{Approaches} & \textbf{Approach} & \textbf{AUC} \\
    \hline
    Harutyunyan et al. \cite{harutyunyan2019multitask} & Time Series & 87.00\% \\
    \hline
    Che et al. \cite{che2018recurrent} & Time Series & 85.29\% \\
    \hline
    Shukla et al. \cite{shukla2021multi} & Time Series & 85.91\% \\
    \hline
    Deznabi  et al. \cite{deznabi2021predicting} & Only Text & 87.50\% \\
    \hline
    Deznabi  et al. \cite{deznabi2021predicting} & Hybrid & 89.90\% \\
    \hline
    Khadanga et al. \cite{khadanga2019using} & Hybrid & 86.50\% \\
    \hline
    Jin et al. \cite{jin2018improving} & Hybrid & 87.53\% \\
    \hline
    Baseline & Only Text & \textbf{86.73\%} \\
    \hline
    \FineEHR & Only Text & \textbf{97.18\%} \\
    \hline
    \end{tabular}%
  \label{tab:compare}%
\end{table}%

In subsection~\ref{subsec:mortality}, we surveyed three mortality prediction approaches on time-series data and clinical notes in EHR. Most of them only presented the AUC result of mortality prediction. Therefore, we only compare their AUC with \FineEHR to demonstrate its improvement. Table~\ref{tab:compare} shows the references, approach, and AUC (presented in the companion paper), as well as the Baseline Approach and \FineEHR using Random Forest Classifier. The Baseline Approach has a lower AUC than most previous works. However, \FineEHR demonstrates a state-of-the-art performance and improves the AUC by nearly 10 percent compared to previous works.

%% file: conclusions.tex
\section{Conclusions}

Clinical text understanding and representation play an essential role in clinical prediction tasks. In this paper, we studied approaches to refine clinical embeddings for better prediction performance downstream. We focus on two representation learning techniques: metric learning and fine-tuning.

We designed a system called \FineEHR to refine text embeddings based on our domain understanding of clinical texts' complexity on various note sources (categories). It inputs raw EHR notes and mortality status, uses Siamese Network to represent raw embedding and adjust distances between note-level embedding pairs, and finally fine-tunes a linear model to combine multiple category-level embeddings into unique admission-level embedding with various weights.

Experiment results show that both refining approaches can improve AUC and AUC-PR compared to a baseline setting. Their combination achieves the highest performance, with an average AUC of 96.04\% and an average AUC-PR of 96.48\% on four famous classifiers. \FineEHR also outperforms all previous approaches, with an AUC improvement of over 10\%.